\documentclass{llncs} 
\usepackage{graphicx}
\usepackage{epstopdf}
\usepackage{subfig}

\usepackage{url}
\urldef{\mailsa}\path|saeed.rad@epfl.ch|    

\begin{document}

\title{A Computer Vision System to Localize and Classify Wastes on the Streets}
\author{Mohammad Saeed Rad\inst{1}, Andreas von Kaenel\inst{2}, Andre Droux\inst{2}, Francois Tieche\inst{3}, Nabil Ouerhani\inst{3}, Haz{\i}m Kemal Ekenel\inst{4}, Jean-Philippe Thiran\inst{1}}
\institute{ 
Ecole Polytechnique Federale de Lausanne, Signal Processing Laboratory 5, Lausanne, Switzerland
\and
Cortexia SA, Chatel-Saint-Denis, Switzerland
\and
Haute Ecole Arc, St-imier, Switzerland
\and
Istanbul Technical University, Istanbul, Turkey\\
\mailsa}
\maketitle
 \vspace{-2mm}
\begin{abstract}
Littering quantification is an important step for improving cleanliness of cities. When human interpretation is too cumbersome or in some cases impossible, an objective index of cleanliness could reduce the littering by awareness actions. In this paper, we present a fully automated computer vision application for littering quantification based on images taken from the streets and sidewalks. We have employed a deep learning based framework to localize and classify different types of wastes. Since there was no waste dataset available, we built our acquisition system mounted on a vehicle. Collected images containing different types of wastes. These images are then annotated for training and benchmarking the developed system. Our results on real case scenarios show accurate detection of littering on variant backgrounds.
\end{abstract}
\section{INTRODUCTION}

Urban littering, defined as the waste products disposed improperly in cities, has recently become a major concern for our modern cities. Major European cities place urban cleanliness as a top priority for the authorities, as it directly impacts the concern and satisfaction of their citizens and the attractiveness of their economy and tourism. At a recent Clean Europe Network summit\footnote{http://www.cleaneuropenetwork.eu/de/measuring-litter/aus/}, the lack of data has been pointed out as one of the major difficulty in addressing properly this environmental issue.

The key to properly manage urban cleanliness is to implement a continuous improvement management system. The measurement of urban litter is mandatory for such a process. Anti-littering organizations such as AVPU\footnote{http://www.avpu.fr/pdf\%20AVPU/formation\%20grille\%20IOP-2014.pdf} and cities worldwide are assessing urban cleanliness by means of human audits. Zurich -ranked third over 83 European cities for the satisfaction of its citizens regarding cleanliness\footnote{http://ec.europa.eu/regional\_policy/sources/docgener/studies/pdf/urban/survey2015\_en.pdf}- is conducting 14’000 audits a year to assess and manage its cleanliness. To provide such a measurement, as an index of cleanliness, a key step is to be able to recognize different types of wastes on urban places, to quantify and to classify them by their type. 

\begin{figure}
   \centering
   \hspace*{\fill}
  \subfloat[]{\includegraphics[height=0.39\textwidth]{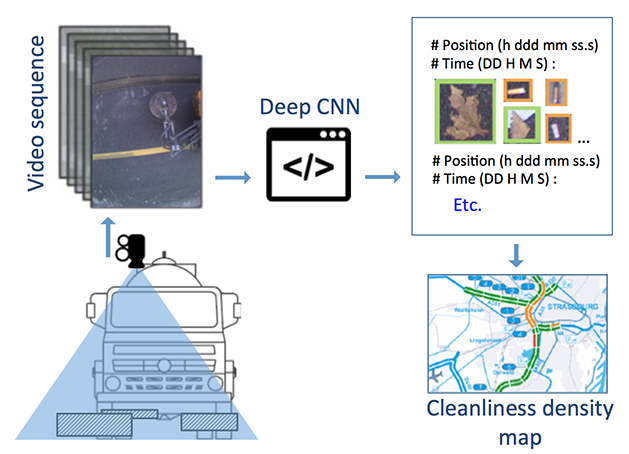}\label{fig_arch}}
  \hfill
  \subfloat[]{\includegraphics[height=0.39\textwidth]{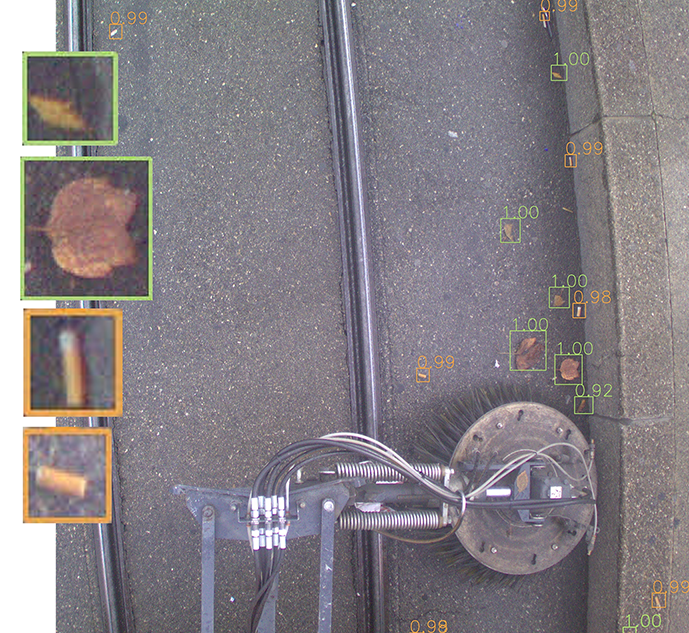}\label{fig_arch_result}}
  \hspace*{\fill}
  \caption{(a) The system overview: The application focuses on detecting different types of urban wastes in RGB images taken by a high-resolution camera, mounted on a vehicle and facing the ground. Its results could be merged in order to produce a waste density map. (b) A visual representation of results obtained by the application: Some detected objects are cropped and highlighted.}
  \label{fig_architecture}
\end{figure}

In this study, we propose and develop a computer vision application based on deep CNN algorithms to localize and classify urban wastes such as bottles, leaves, etc.  in an automated manner in RGB images. This measurement is realized by an image acquisition system consisting of a high-resolution camera, mounted on the top of a vehicle, facing the ground. The front surface of the vehicle are covered by the camera view. The system must be able to detect the smallest defined waste -a cigarette butt, seen from a camera placed at a height of two to three meters. The output of this application is a geo-localized density of different categories of urban wastes. An overview of the system is shown in Fig.~\ref{fig_architecture}.

The remainder of this paper is structured as follows. Section \ref{section_relatedwork} presents related works. Then, in Section \ref{section_methodology}, the deep neural network used for detection -including its implementation details- is explained. In Section \ref{section_dataset}, we present the data collection setup and use it to obtain a waste dataset. In Section \ref{section_results} we test our application on a real case scenario and present results. Finally, Section \ref{section_summary} summarizes our work and discusses about future work.

\section{RELATED WORK}\label{section_relatedwork}
Different methodologies have been developed worldwide to obtain an index of cleanliness for a city. These approaches are mostly focused on human interpretations of cleanliness. However, an automated approach has not yet been developed.

The closest work to this study is a trash related project designed to coarsely segment a pile of garbage in an image \cite{cit_garbnet}. They also provide an Android application, which allows citizens to track and report garbage in their neighborhoods. Bing Image Search API\footnote{https://www.microsoft.com/cognitive-services/en-us/bing-image-search-api} was used to create their dataset. They have labeled images as containing garbage or not. The authors utilize a pre-trained AlexNet \cite{cit_alexnet} model and obtain 83.96\% of sensitivity with 90.06\% specificity. Their approach focuses on segmenting a pile of garbage in an image and provides no details about types of wastes in that segment.

There exist approaches that classify garbage into recycling categories; \cite{cit_tiar} proposes an automated recognition system using deep learning algorithm which classifies objects as biodegradable and non-biodegradable. They propose a model and have its implementation done in Caffe \cite{cit_caffe}. However, there are no experimental results presented. In \cite{cit_colombian}, they propose a system to classify waste in high schools. They design a box containing a camera inside it. In order to do the classification, objects are required to be placed inside the box. Their image processing module is based on finding correlation between the image of the object in the box and 50 different images, then choosing the best one as the right category. The developed system classifies three kinds of waste: PET bottle, soda cans and cartoon box, with a classification performance over 70\%.

An automatic waste sorting approach is presented in \cite{cit_sakr}. They use two different methods: Convolution Neural Networks and Support Vector Machines. Their input is $256 \times 256$ pixel resolution image of the waste. For their CNN architecture, they use AlexNet\cite{cit_alexnet} model. Their SVM utilizes a bag of features obtained by passing a $8 \times 8$ window over the whole image. Each algorithm creates a different classifier that separates waste into three main categories: plastic, paper, and metal. They achieved a classification accuracy of 94.8\% with SVM, while CNN had an accuracy of 83\%. As they have mentioned in their paper, the main reason of not having better results with CNN is the insufficient number of images in their training set. Their approach focuses on classifying a specific object and not to localize it from a far distance. 

\section{METHODOLOGY}\label{section_methodology}

In this section, we describe each step of our approach.  We explain how we localize and classify the wastes on given input images, then we discuss about the implementation details.

\subsection{Waste Localization and Classification}

The proposed system must take care of two main tasks: The first task is to localize all objects in the image. The second task is to classify all detected objects on their right littering category. In this section, all tasks are addressed using a single framework and a shared feature learning base.

The fact that CNNs are trained end-to-end, from raw pixels to final classes, makes them much more advantageous for many tasks than manually designing a suitable feature extractor. Our approach is similar to OverFeat model \cite{cit_overfeat} which proposes a multi-scale deep learning approach that can be used for classification, localization and detection. We replace its classification architecture by GoogLeNet \cite{cit_googlenet}.  For localization, as OverFeat put forward, starting from the classification-trained network, the classifier layers are replaced by a regression network and trained to predict object bounding boxes at each spatial location and scale. Then the regression predictions with the classification results are combined at each location to obtain detection results. Object bounding box predictions are generated by running the classifier and regressor networks for all locations and scales. Considering that these two networks are sharing the same feature extraction layers; after computing the classification net, only the final regression layers must be recomputed. The final output layer of regression network has 4 units which correspond to coordinates for the bounding box of the detected object.

We use OverFeat-GoogLeNet model presented in \cite{cit_russell}. The original version of OverFeat relies on image representation based on AlexNet \cite{cit_alexnet}. In \cite{cit_russell}, they were able to directly substitute the GoogLeNet architecture into the OverFeat model and denoted the new model as OverFeat-GoogLeNet. They show that Overfeat-GoogLeNet performs significantly better than OverFeat-AlexNet. GoogLeNet is initially trained on 1.2 million images for 1000-classes object recognition. Overfeat-GoogLeNet uses expressive image features from GoogLeNet that in our implementation are fine-tuned as part of our system. The size of the input layer is fixed to $640 \times 480$ pixels. The model is constructed to encode the input image into a $15 \times 20$ grid where each cell contains 1024-dimensional top level GoogLeNet features and has a receptive field of size $139 \times 139$. Cells are trained to produce the set of all bounding boxes intersecting the central $32 \times 32$ region. The convolutional layers are followed by two fully connected layers containing 3092 and 4096 neurons, respectively. At the end, the output layer contains 25 neurons corresponding to different categories of waste.

\subsection{Implementation}\label{section_implementation}

An open source implementation of OverFeat on Tensorflow \cite{cit_tensorflow} was used as a starting point. Then, some modifications were done to perform multi-classification. 
The image of a cigarette butt must contain at least same number of pixels as the smallest possible bounding box for the network. To fulfill this last criterion, and also regarding to the height of the camera, the resolution is fixed to $1920 \times 1480$ pixels. During training, these images occupy a considerable amount of memory while loading their batches. Due to this and the challenge of having a cheaper system capable of processing and detecting wastes onboard on an embedded system, we decided to pass a $640 \times 480$ pixels sliding window with an overlapping factor over the input image and keep the network input size same as the window size. The final result is produced by converting the detection coordinates with respect to initial full image. Detections within the same category are merged in case of having an overlapping factor of more than 60\%. The model is fine-tuned on Tensorflow \cite{cit_tensorflow} using Nvidia K40 GPUs for 350,000 iterations with a batch size of 16. Validation is performed every 2,000 iterations.

\section{THE DATASET}\label{section_dataset}

Convolutional Neural Networks have lots of advantages over methods requiring to design a suitable feature extractor. However, one of their drawbacks is the need for a large amount of labeled training samples.

There is no waste image dataset currently available, which differentiates different types of litters/wastes. Our initial idea was to gather a diverse set of images, for example using image search by entering the category names as the keywords or using ImageNet \cite{cit_imagenet}, to train our system. However, the final decision was to not use them for training as their conditions like camera view, illumination, etc. were too different from what our system captures. To collect our own dataset, we have built our own acquisition system, mounted on a vehicle and drove several hours in Geneva area, Switzerland. We have obtained 18,676 images. To avoid overlapping between training images we have decreased number of images from 2 to 0.4 frame per second and among them we have annotated 469 full images, which corresponds to 4338, $640 \times 480$ pixel resolution images. Because of the time and season of our acquisition process, most of wastes found in images were leaves and cigarette butts.

\subsection{Categories}\label{section_categories}
Another important step is to define what the waste is and needs to be considered for a cleanliness measure, and also how the categories should be defined in order to cover most of litters. 
Different organisations use different waste classifications. The OFEV\footnote{http://www.bafu.admin.ch/publikationen/publikation/01604/index.html?lang=fr} approach, for example, does not take into account gums or excrement, which nevertheless play an important role in the perception of cleanliness and urban pollution. To give an example, in Roma, 5.54 million gums are discarded every year that take about 5 years to degrade. In this work, after some discussion with different cities we have decided to classify different wastes into one of the 25 general categories. Here we mention some important ones: 1. Beverage and meal packages, 2. Cigarettes and derivatives, 3. Leaves, 4. Newspapers and papers, 5. Vegetable waste, etc.

\subsection{Setup}\label{section_setup}
\begin{figure}
\centering
 \vspace{-1mm}
\includegraphics[scale=.22]{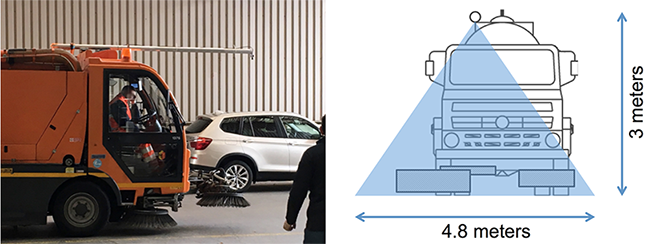}
\caption{Our dataset is obtained by a high resolution camera mounted on top of a street sweeper and having a flat view of the ground. The camera is in a distance of approximately three meters from the ground.}
\label{fig_streetsweeper}
\end{figure}

We equipped an automatic street sweeper car with a camera and an embedded system to obtain and store our dataset. As it is shown in Fig. \ref{fig_streetsweeper}, the camera was installed on a metallic arm, on top and coming out of the vehicle, having a flat view of the ground. The camera has a rolling shutter with a 1/2.3 inch CMOS sensor and 4K resolution, however after tuning the input image size of the network, the camera was configured to an output of $1920 \times 1480$ pixel resolution images. The camera was set to get two frames per second and the average speed of the vehicle was twelve kilometers per hour. 

\subsection{Annotations}
\begin{figure}
\vspace{-5mm}
\centering
\includegraphics[scale=1.2]{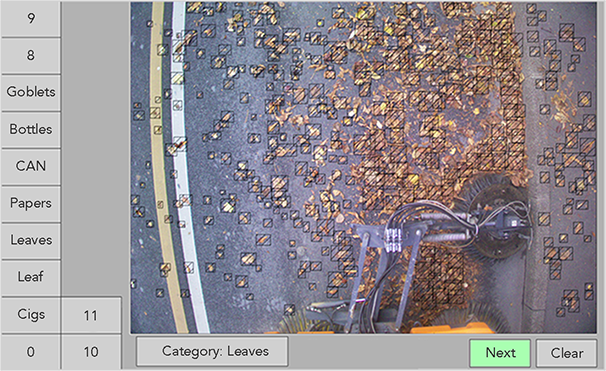}
\caption{A screenshot of our annotation tool. This image shows existing challenges to label collected images.}
\label{fig_gui}
\vspace{-1mm}
\end{figure}

\begin{figure}
\vspace{-7mm}
\centering
\includegraphics[scale=.35]{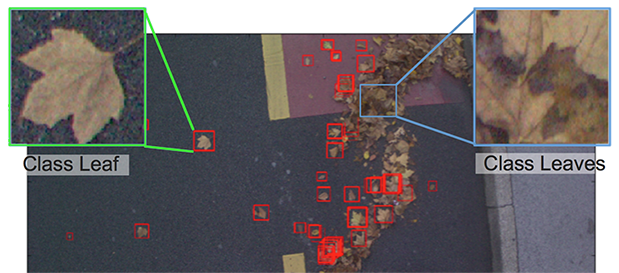}
\caption{The red detections are done by the network only trained with a single category for leaves. A significant improvement was seen when two different classes were defined for leaves:  1- Class leaf, 2- Class leaves.}
\label{fig_leaves}
\end{figure}

Similar to other object recognition problems, our model also requires considerable amount of labeled training samples. We have developed an annotation tool to label a sequence of images by putting a bounding box around each waste and assigning an integer number to it showing its class number. A screenshot of this tool is shown in Fig. \ref{fig_gui}. This approach is based on the hypothesis that each object is well-separated, countable and has its particular shape, which is not the case for all categories. For example during autumn, the ground is covered by leaves where each individual leaf will not appear the same way that it appears alone. A significant improvement was observed in the correct classification accuracy once two different classes were introduced for leaves: a class for single leaves and another class for piles of leaves. However, for the cleanliness measurement both classes are considered as one category. This approach helped the network to have a better generalization for each type, separately. An example of these two classes is shown in Fig. \ref{fig_leaves}.

\section{RESULTS AND DISCUSSION}\label{section_results}
The proposed application was validated using a test set consisting of 62 non-overlapping full-size images collected from our setup, equal to 558 $640 \times 480$ images that are fed to the network. Rectangular ground-truth bounding boxes were defined on each image. In total, they consist of 69 cigarettes, 958 leaves and 394 bounding boxes on piles of leaves. Although other types of waste had been annotated and were used during training, they were not considered for the evaluation. Their number was not sufficient and could not provide a reliable training/testing. For example, in total we have: 8 bottles, 5 cans, 6 goblets in training set.

We have reached to process images at 2 frames ($1920 \times 1480$) per seconds. This could  be interpreted as: with a camera mounted at a height of 3 meters, we can detect a cigarette butt with a speed up to 12 kilometers per hour (this number of frame per second enables us to have 15\% of overlap between two consecutive images). Both training and testing processes were done on a Nvidia K40 GPU.

\subsection{Precision-recall analysis}
To evaluate the performance of the proposed application in a quantitative manner, a precision-recall analysis was performed \cite{cit_pascal}. The precision ($P$) and recall ($R$) rates of the system are simply defined as: $P = CD/ (CD + FP)$ and $R = CD/N$ where $CD$, $FP$ and $N$ are the total number of correct detections, false positive and ground truth objects respectively.

\begin{figure}
\vspace{-5mm}
  \centering
  \hspace*{\fill}
  \subfloat[]{\includegraphics[width=0.38\textwidth]{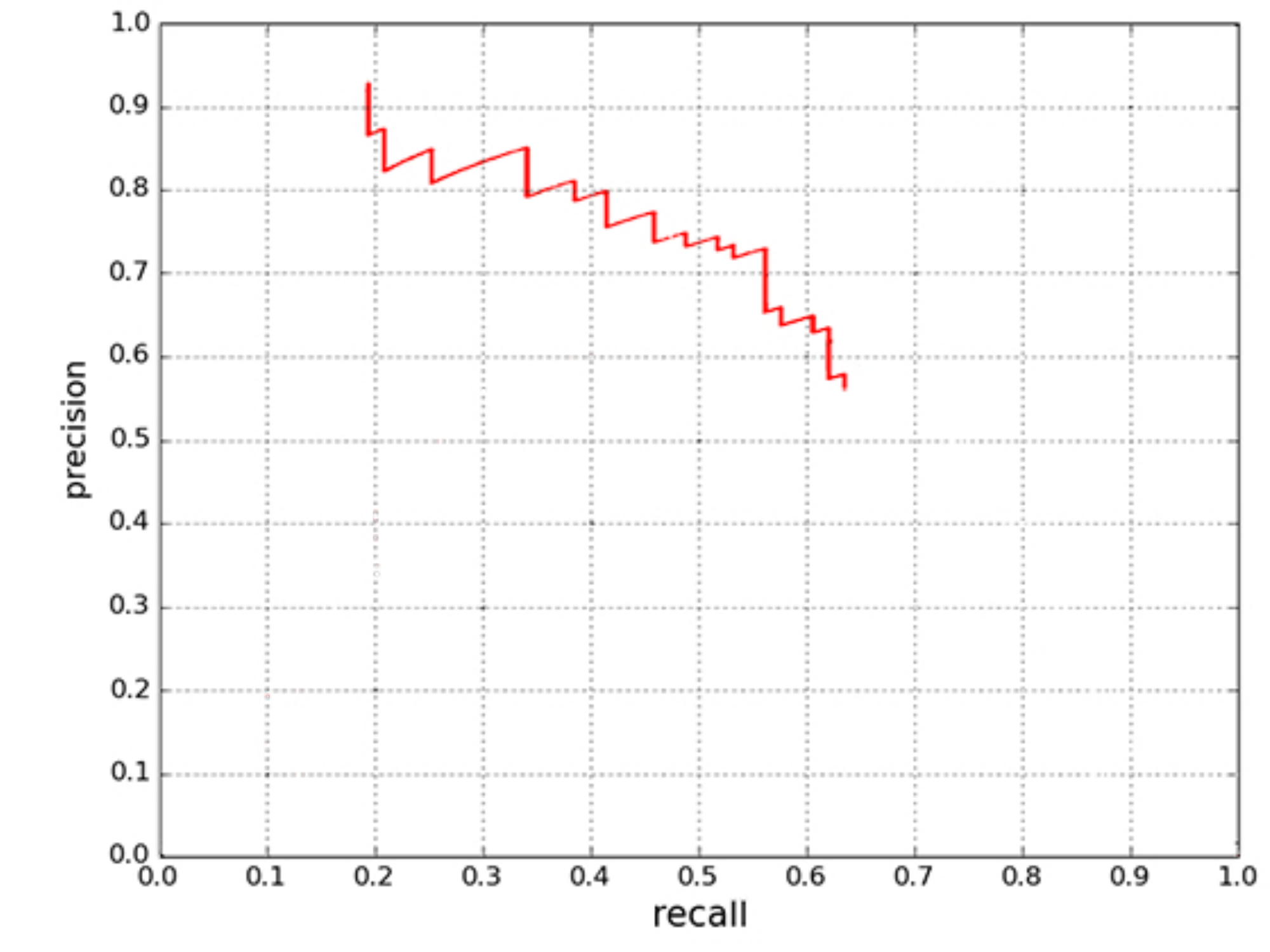}\label{fig_pr_cigs}}
  \hfill
  \subfloat[]{\includegraphics[width=0.38\textwidth]{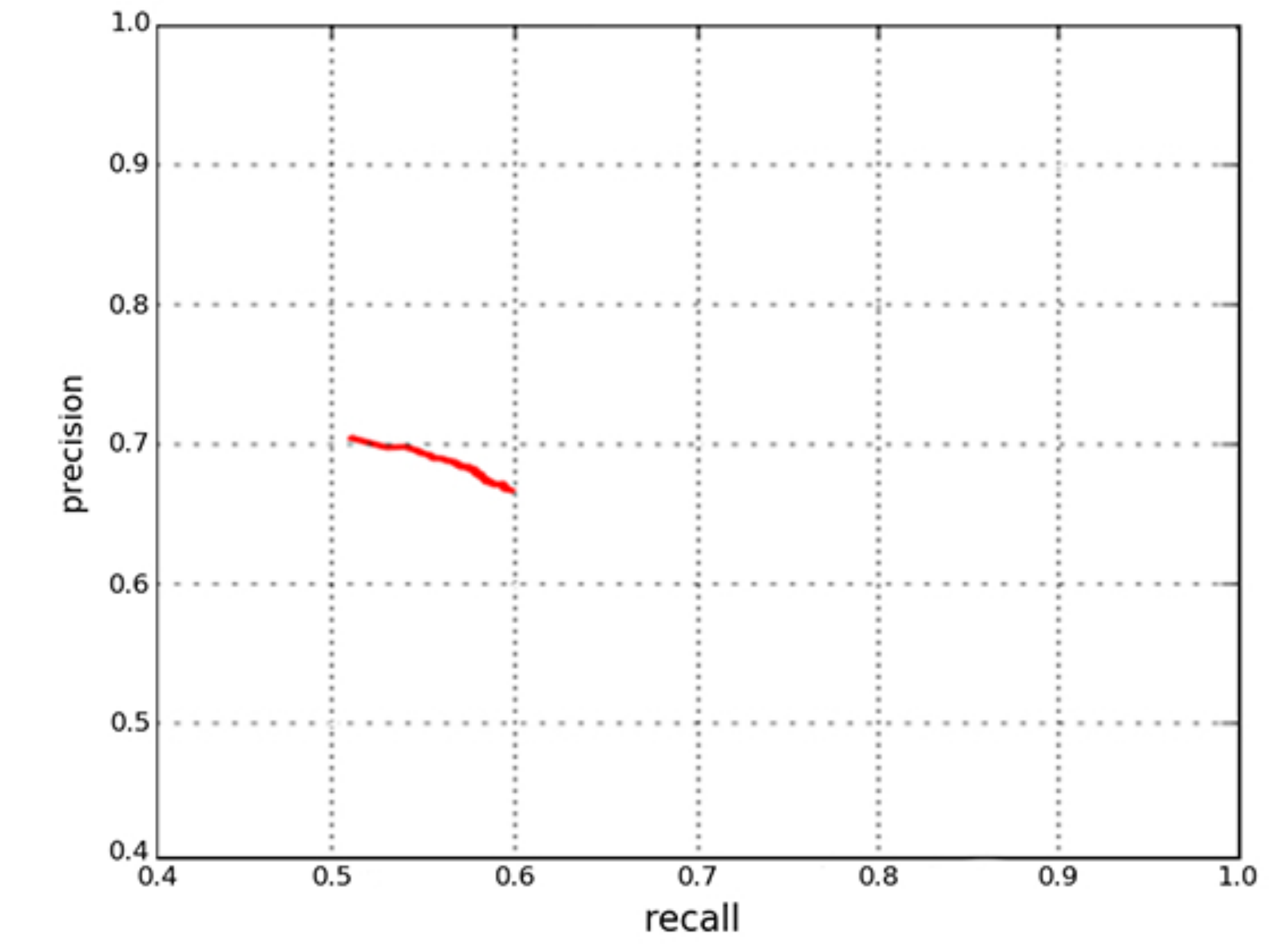}\label{fig_pr_leaves}}
  \hspace*{\fill}
  \caption{Precision-recall curves for: (a) Cigarettes class, (b) Leaves class }
  \label{fig_precision_recall}
  \vspace{-5mm}
\end{figure}

In order to calculate these parameters, first, each detection needs to be  labeled as either correct detection or false positive by reference to the ground truth. For cigarette butts category, a detection is marked as correct when the overlap between its detected bounding boxes and the corresponding ground truth is at least 50\%. In Fig.~\ref{fig_precision_recall}.(a) the precision and recall of cigarette butts is illustrated. Different values for $P$ and $R$ are obtained by varying a threshold on final detection score for this category. We have reached 63.2\% of precision while having 61.02\% of recall for the cigarette butts class.

This method of defining correct detection and false positive could pose a problem while evaluating the application for some categories like leaves. Let's imagine a scene covered by leaves. As explained previously, our ground-truth is defined by different overlapping bounding boxes, with different sizes, on some random position, covering leaves. In this case, the algorithm would correctly return different detections on top of leaves' regions, but not exactly the same position that was defined in the ground-truth. To avoid this issue, only for this category, a binary image of detection/ground-truth was produced for each image. Pixels set to 0 indicate background and pixels set to 1 show ground-truth/detection. Comparing these two binary images pixel by pixel gives $CD$ and $FP$ parameters. Fig.~\ref{fig_precision_recall}.(b) shows precision-recall curve for leaves category. We have obtained 77.35\% of precision while having 60\% of recall for the leaves class.

Although the quantitative results may not seem to be too high, it should also be taken into account that the system is designed to localize very small objects such as cigarette butts in relatively large images, covering five meters of a street. Considering this challenge, these results are promising for waste localization and classification even if they are seen from a distance.

\subsection{Qualitative assessment}
\begin{figure}
  \centering

  \subfloat{\includegraphics[height=0.28\textwidth]{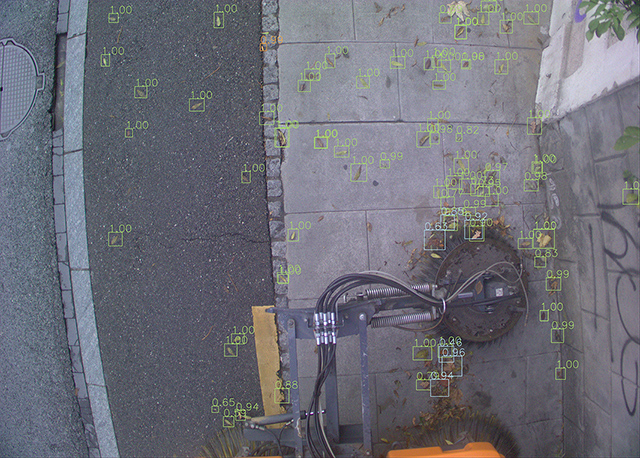}\label{fig:f1}}
  \hspace{0em}
  \subfloat{\includegraphics[height=0.28\textwidth]{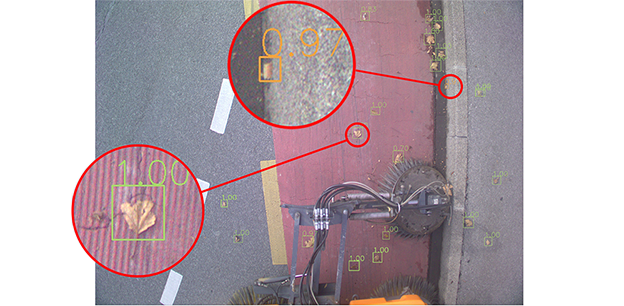}\label{fig:f2}}

  \subfloat{\includegraphics[height=0.28\textwidth]{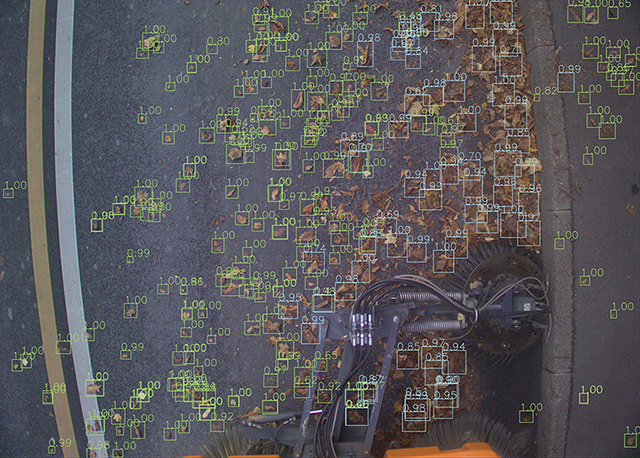}\label{fig:f3}}
  \hspace{0em}
  \subfloat{\includegraphics[height=0.28\textwidth]{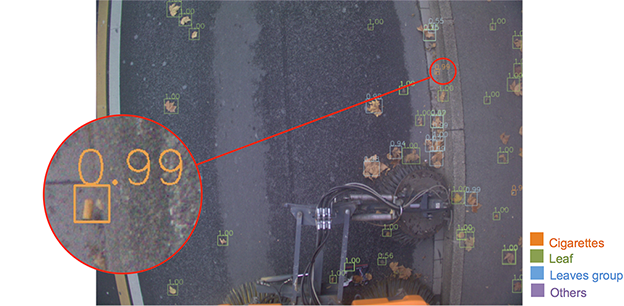}\label{fig:f4}}
  
  \caption{Example of leaves and cigarettes detections. Orange boxes correspond to cigarettes, green boxes to leaves and blue boxes to piles of leaves. }
  \label{fig_results}
\end{figure}

\begin{figure}

  \centering

  \subfloat[]{\includegraphics[height=0.28\textwidth]{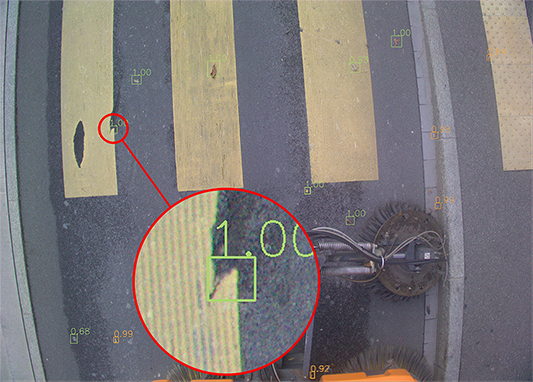}\label{fig_res_bad1}}
  \hspace{0em}
  \subfloat[]{\includegraphics[height=0.28\textwidth]{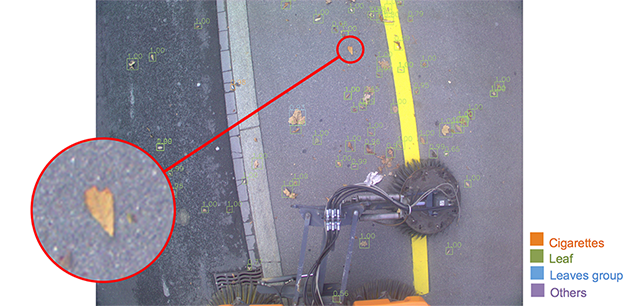}\label{fig_res_bad2}}

  \caption{Example of: (a) A false detection (b) A missed detection.}
  \label{fig_results_bad}
  
\end{figure}

Some localization and classification results obtained on sample representative images
are shown in Fig. \ref{fig_results}. The proposed approach performs well for small objects like a cigarette butt from a three meters height on a clear background as well as in backgrounds crowded by other types of waste. Also, the method is able to detect
multiple/overlapped wastes. It should be noted that some leaves/cigarettes were missed on some images, which could be due to our limited training-set. Examples of a false positive detection and a missed detection are shown in Fig.~\ref{fig_results_bad}.

\section{SUMMARY AND FUTURE WORK}\label{section_summary}
In this paper, a novel application for measuring cleanliness of a place, using a deep learning framework was proposed. The application localizes and classifies wastes in RGB images taken by a camera facing ground from three meters of height. Since there was no waste dataset available, we used our proposed acquisition setup to obtain images. We have also developed an annotation tool to label objects in our dataset for 25 different types of waste. Experimental results on a real case scenario -on a test-set obtained by our proposed acquisition setup- show promising performance on variant backgrounds. 

As a future work, our dataset could be expanded by adding more images, especially for categories different than cigarette butts and leaves, to be able to detect all existing classes of wastes, and also to increase the accuracy of the current system. 

\section*{ACKNOWLEDGMENT}
The authors would like to thank Mr. Niels Michel, Manager of Dialog \& Service at City of Zurich for sharing his in-depth experience on cleanliness measurement thus significantly contributing to this project. \\

The final publication is available at link.springer.com via  \url{http://dx.doi.org/10.1007/978-3-319-68345-4_18}.

\end{document}